\documentclass[letterpaper, 10 pt, conference]{ieeeconf}  
\IEEEoverridecommandlockouts                              
\usepackage{amsmath}
\usepackage{amssymb}
\usepackage{algorithm}
\usepackage{algpseudocode}
\usepackage{mathtools}
\usepackage{bm}
\usepackage{graphicx}
\usepackage{lipsum}
\usepackage{authblk}

\newcommand\drawiid{\stackrel{\mathclap{\normalfont\mbox{\textit{iid}}}}{\sim}}

\title{\LARGE \bf
Sequential Bayesian Optimisation as a POMDP \\ 
for Environment Monitoring with UAVs}

\author{Philippe Morere, Roman Marchant and Fabio Ramos
\thanks{Philippe Morere, Roman Marchant and Fabio Ramos are with the School of Information Technologies, University of Sydney, NSW 2006 Australia.}
\thanks{{\tt\small \{philippe.morere, r.marchant, fabio.ramos\} @sydney.edu.au}}
}

\begin{document}

\maketitle
\thispagestyle{empty}
\pagestyle{empty}

\begin{abstract}
\textit{Bayesian Optimisation} has gained much popularity lately, as a global optimisation technique for functions that are expensive to evaluate or unknown {\em a priori}. 
While classical BO focuses on \textit{where} to gather an observation next, it does not take into account practical constraints for a robotic system such as where it is physically possible to gather samples from, nor the sequential nature of the problem while executing a trajectory. In field robotics and other real-life situations, physical and trajectory constraints are inherent problems.
This paper addresses these issues by formulating Bayesian Optimisation for continuous trajectories within a \textit{Partially Observable Markov Decision Process} (POMDP) framework. The resulting POMDP is solved using \textit{Monte-Carlo Tree Search} (MCTS), which we adapt to using a reward function balancing exploration and exploitation. Experiments on monitoring a spatial phenomenon with a UAV illustrate how our BO-POMDP algorithm outperforms competing techniques.
\end{abstract}

\section{Introduction}
Path planning under uncertainty is central to environmental monitoring with mobile robots.
The problem can be formalised as solving a \textit{partially observable Markov decision process} (POMDP), a popular and mathematically principled framework.
POMDPs allow for planning when the state of the world is hidden or stochastic. This is achieved by maintaining a probability distribution, called belief, over the set of possible states the world could be in. The transition and reward dynamics of the world are also defined as probability distributions. They can be used by POMDP solvers to simulate the execution of sequences of actions and compute their associated rewards. POMDP based planners can efficiently compute the best action, looking several steps ahead, while dealing with stochastic transition dynamics and sensor noise.

The task of picking trajectories to maximise information gain is known as informative path planning, and is a fundamental concept of monitoring. Reasoning over trajectories instead of over destinations was shown to increase information gain \cite{marchant2014bayesian}.

While many monitoring applications use informative path planning, most rely on myopic approaches. Myopic techniques naively define the best action as the one with highest immediate reward. By failing to account for future potential rewards, myopic algorithms often yield suboptimal behaviour.
Classic non-myopic planners compute the whole trajectory sequence offline. While these planners can generate optimal policies when dealing with the most engineered cases, they fail on real-life scenarios or when little or no data is known a priori. Their main shortcoming is their inability to improve the sequence of trajectories with the information gathered.
On the other end of the spectrum lies online planning: after each trajectory, robots recompute the best action by simulating numerous sequences of actions, thereby sampling the potentially infinite space of possible scenarios. Finding the best action is a non-trivial maximisation problem, especially when robots evolve in stochastic and noisy environments.
\\

This paper presents BO-POMDP, a POMDP based informative path planning algorithm based on \cite{marchantRS14}, suitable for monitoring real-life environmental phenomena. BO-POMDP uses a Gaussian process to maintain a belief over the studied phenomenon, thus including uncertainty information when reasoning. By redefining the POMDP reward function as a balance between belief uncertainty and gradient, we are able to induce an exploration-exploitation behaviour. Encouraging the robot to exploit high-gradient areas is paramount to achieving high quality monitoring, as reconstructed maps usually suffer from higher errors in complexly varying areas.
BO-POMDP simulates several steps ahead and recomputes the best action after each trajectory finishes, making it a non-myopic online planner. The inherent complexity of online POMDP planners is dealt with by using an adapted version of \textit{Monte-Carlo tree search} (MCTS) to approximate the aforementioned maximisation problem.

Our contributions are twofold. We first present a myopic planner and our BO-POMDP algorithm, both improving on \cite{marchantRS14} by redefining a reward function trading off exploration and exploitation of gradients. We show in simulation that such reward function yields superior results to the method our work is based on. The comparison between our myopic and non-myopic algorithms suggests planning with higher horizons improves monitoring behaviour.
Secondly, we display the first practical results of the BO-POMDP formulation, using a cheap quad-copter to carry out real-life monitoring. We show the method presented in this paper can be used to successfully map terrains, only using limited sensing capabilities.

The remainder of this paper is organised as follows.
Section \ref{sec:RelatedWork} reviews the existing literature on informative path planning and POMDP planners.
Section \ref{sec:Background} gives an introduction to Bayesian optimisation and POMDPs. Section \ref{sec:Method} presents a myopic solution to informative path planning, then extending to the non-myopic case. Section \ref{sec:ExperimentalResults} displays simulation and real-life experimental results. The paper concludes with Section \ref{sec:Conclusion}.

\section{Related work}
\label{sec:RelatedWork}
Path planning and decision making under uncertainty have been studied by many researchers. Even though much theoretical work exists on the topic, few advanced methods were successfully applied to real-world robots.

Basic planning methods which only consider the next destination are qualified of \textit{myopic}. Myopic planners are efficient to solve basic problems, but their lack of anticipation makes them fail on more complicated tasks. \textit{Bayesian optimisation} (BO) is a perfect framework for myopic decision making. It was successfully used in \cite{marchant2012bayesian} to pick the next best destination a robot should move to. The in-built exploration-exploitation balance of BO yields good monitoring behaviour.
Exploiting the information robots gather along their trajectories to improve planning is referred as \textit{informative path planning} in the literature. Informative path planning enables robots to take more complicated and more informative trajectories, by reasoning over the route instead of the destination. For example, the technique described in \cite{marchant2014bayesian} defines the best trajectory as the one maximising the sum of rewards given for travelling along it. The main shortcoming of this type of method is it neglects long-term consequences when taking decisions. Indeed, not planning further than a single action ahead can result in sub-optimal situations (eg. robot facing a wall, requiring a sharp turn, trapped in a low-reward area).
\\

Several frameworks for non-myopic planning were developed over the last decades. One of the best defined and most successful ones is the \textit{partially observable Markov decision process} (POMDP), which allows the robot's current state to be unobserved. POMDPs allow one to simulate taking sequences of $n$ actions, and computing the overall reward of executing such sequence. The action leading to the highest sequence of rewards is selected for execution. More theory is given in section \ref{sec:BackgroundPOMDP}.

Over the years, many POMDP solvers were developed. Their high complexity favours offline solutions such as \cite{pineau2003point}. However, because the robot has no a priori information of the studied phenomenon, offline solutions cannot be used in our set-up.

Several online POMDP solvers exist, and often tackle high complexity by relying on sampling methods \cite{he2010puma, kurniawati2016online, ross2008online, silver2010monte, somani2013despot, kurniawati2008sarsop}. Some of them were extended to the case of continuous state and observation spaces \cite{kurniawati2016online, silver2010monte, somani2013despot}, and to continuous action spaces \cite{seiler2015online}. These POMDP solvers deal with large and continuous state and observation spaces by sampling them to build more compact representations. This approximation step does not satisfy our needs, as maps built from sampled spaces suffer from poor resolution in areas of interest, or incur tremendous costs on the planner when very high sampling is used. Furthermore, we wish to have access to the uncertainty of the state space to easily determine areas to explore.
\\

Another method is proposed in \cite{marchantRS14}, based on Monte-Carlo Tree search for POMDPs \cite{silver2010monte} and using a Gaussian process (GP) to maintain a belief over a continuous state space. Modelling the belief with a GP is an elegant way to tackle continuous states and observations while keeping generality with the type of functions the belief can represent \cite{rasmussen2006gaussian}. Also, GPs naturally provide uncertainty information, a core feature for balancing exploration-exploitation.
Defining the POMDP reward function as a BO acquisition function, \cite{marchantRS14} takes advantages from both techniques to achieve non-myopic informative path planning.
While most of the work we reviewed so far is theoretical, decision making under uncertainty has also been applied to robots.
\\

Planning for fault inspection is achieved using the travel salesman problem on a submarine in \cite{hollinger2012uncertainty}. The underwater glider in \cite{witt2008go} relies on simulated annealing and swarm optimisation to find energy-optimal paths in strong currents. Informative path planning is carried out in \cite{binney2012branch} by using a variant of branch and bound. These methods minimise the overall uncertainty of the underlying model, therefore not fitting our problem definition in which different levels of certainty must be achieved across the state space.
Finally, a method balancing global uncertainty reduction and re-sampling of areas of interest is tested on a ground robot in \cite{marchant2014bayesian}. The major drawback of this method however is its inability to plan more than one step ahead. This property makes the planning algorithm converge to greedy suboptimal paths.

We propose to build on the work of \cite{marchantRS14} and \cite{marchant2014bayesian} to formulate a non-myopic planning algorithm for real-life environmental monitoring. We first modify the reward function defined in \cite{marchantRS14} to achieve a monitoring behaviour that exploits high-gradient areas. We then implement and adapt the algorithm to a real application of UAV terrain mapping.

\section{Background}
\label{sec:Background}
We start with a brief description of Bayesian Optimisation, and then present the classic POMDP formulation.
\subsection{Bayesian Optimisation}
Bayesian optimisation is a technique aiming to find the optimum $\hat{x} \in \mathbb{R}^D$ of an objective function $f : \mathbb{R}^D \rightarrow \mathbb{R}$ by gathering noisy observations from it. Formally,
\begin{equation}
\label{eq:bo}
\hat{x} = \arg\max_x f(x)
\end{equation}
Noisy observations are assumed to result from an additive Gaussian noise on the function evaluation. The $i$th observation is defined as $y_i = f(x_i) + \epsilon$, where $\epsilon \drawiid \mathcal{N}(0, \sigma_n^2)$ is the noise associated with each independent observation. More theory on BO and GP can be found in \cite{brochu2010tutorial} and \cite{rasmussen2006gaussian}.
Most implementations of BO use Gaussian processes to model the objective function $f$. The GP model is updated with the data couple $(x_i, y_i)$ every time a noisy observation $y_i$ is made at location $x_i$. The search of where to get an observation next is guided by an acquisition function $h(x)$. At each iteration in the BO algorithm, the location at which to evaluate $f$ is determined by finding $\arg\max_x h(x)$, therefore reporting the maximisation problem from $f$ to $h$. $h$ is much easier to optimise with traditional techniques such as DIRECT \cite{jones1993lipschitzian} and, contrary to $f$, is cheap to evaluate. An implementation of BO is detailed in Algorithm \ref{alg:BO}.
\begin{algorithm}[H]
\caption{Bayesian Optimisation} 
\label{alg:BO} 
\begin{algorithmic}[1]
\State Let $x_t$ be the sampling point at iteration $t$.
\State Let $D = \{(x_1, y_1), ..., (x_n, y_n)\}$ be the data.
\State Let $h$ be an acquisition function.
\For{$t = 1, 2, 3, ...$}
\State Find $x_t = argmax_x h(x|D_{1:t-1})$.
\State Sample the objective function $y_t = f(x_t) + \epsilon_t$.
\State Augment the data $D_{1:t} = \{D_{1:t-1}, (x_t, y_t)\}$.
\State Recompute the GP model with $D_{1:t}$.
\EndFor
\end{algorithmic}
\end{algorithm}
\subsection{Partially Observable Markov Decision Processes}
\label{sec:BackgroundPOMDP}
Partially observable Markov decision processes (POMDP) are a framework for decision making under uncertainty \cite{smallwood1973optimal}. Unlike in Markov decision processes, the agent cannot directly observe its current state. It must instead act relying on a belief of the underlying state, built from observations. It is therefore important for the agent to maintain a probability distribution over the set of possible states.
A POMDP is fully defined by the tuple $<S,A,T,R,\Omega,O,\gamma>$, with:
\begin{list}{-}{}
\item S: Set of states $\{ s_1, s_2, ..., s_n\}$.
\item A: Set of actions $\{a_1, a_2, ..., a_m\}$.
\item T: $S\times A\times S \rightarrow [0,1]$ is a transition function interpreted as the probability to transition to state $s'$ when executing action $a$ in state $s$, i.e. $T(s,a,s')=p(s'|s,a)$.
\item R: $S\times A\rightarrow \mathbb{R}$ is a reward function defining the reward of executing action $a$ in state $s$, i.e. $R(s,a)$.
\item $\Omega$: Set of observations $\{o_1, o_2, ..., o_l\}$.
\item O:  $S\times A\times \Omega \rightarrow [0,1]$ is an observation function that represents the probability of observing $o$ when action $a$ was executed and led to state $s$, i.e. $O(o,a,s)=p(o|a,s)$.
\item $\gamma \in [0,1]$ is the discount factor.
\end{list}
POMDPs rely on the Markov assumption, which states the distribution on future states only depends upon the current state. Hence it is not necessary to keep track of an observation history as \textit{all} information at time $t$ is assumed to be embedded in the current belief state $b_t(s)$.

Solving a POMDP is equivalent to finding the optimal policy $\pi^*: \Omega \rightarrow A$. The optimal policy is defined as the one maximising the expected infinite sum of discounted rewards $r_t$ starting from belief state $b_0$. More formally,
\begin{equation}
\label{eq:policy}
\pi^* = \arg \max_\pi E[\sum_{t=0}^{\infty} \gamma^t r_t^\pi | b_o]
\end{equation}
where $r_t^\pi$ is the reward given for following policy $\pi$ at time t.
\\

Numerous methods to solve POMDPs were proposed over the years. In this context, we are interested in planning, the case in which the transition and reward functions are known. In the literature, planning in POMDPs was addressed with numerous techniques \cite{ross2008online}, such as Value iteration \cite{kurniawati2016online}, heuristic search \cite{somani2013despot, silver2010monte}, and branch-and-bound pruning \cite{kurniawati2008sarsop}. The method we present in this paper is based on a tree search algorithm.

\section{BO-POMDPs}
\label{sec:Method}
We wish to solve the problem of finding the \textit{best} sequence of trajectories along which to gather samples. The value of a trajectory depends on the quantity of information gathered along it and how much it reduces the uncertainty of the global belief on the monitored phenomenon. This value is captured by a metric that balances exploration and exploitation. In the myopic case, in which the planner only looks one step ahead, the metric is computed for a single trajectory. In the non-myopic case however, the value of a trajectory should take into account all expected future samples. Ideally, it should be computed with \textit{all} future samples over an infinite horizon. In practice however, the horizon (or look-ahead) needs to be finite for tractability reasons.
The extreme myopic case of a 1-look-ahead can be tackled by using Bayesian optimisation with an acquisition function balancing exploitation of past high-reward actions and exploration of unseen ones. This approach yields acceptable results for simple problems and is presented in the following section.

\subsection{A myopic continuous planning solution}
\label{sec:myopicSolution}
In this section, we describe a naive solution to the UAV monitoring problem. The approach we present is myopic, but takes advantage of the continuity of trajectories when planning. Most approaches rely on picking a destination at which to gather samples next, then building a trajectory between the robot's pose and the destination that respects trajectory constraints. While this simple approach works well in practice, it neglects the information gathered along the trajectory, making the method sub-optimal. Furthermore, most sensors have a high sampling frequency when compared to the frequency at which planning is carried. Making the most out of the samples is essential to building a good planning algorithm.
To solve this problem, we propose to include the information gathered along the trajectory in the planning process. We define the value $r$ of a trajectory $\mathcal{T}(\Theta, \textbf{p})$ defined by parameters $\Theta$ and starting from pose $\textbf{p}$ as:
\begin{equation}
\label{eq:traj_val}
r(\Theta,\textbf{p}) = \int_0^1 h(\mathcal{T}(\Theta, \textbf{p})|_{u=t})dt
\end{equation}
where $h$ is an acquisition function yielding the desired behaviour.
In practice, the integral must be replaced by a discrete sum, in which the number of points to sum depends on the sensor frequency. The Upper Confidence Bound acquisition function \cite{cox1997sdo} was selected for its intrinsic balance between exploration and exploitation, then modified to exploit areas of high gradient. Equation \ref{eq:traj_val} then becomes:
\begin{equation}
\label{eq:traj_val_belief}
r(\Theta, \textbf{p}, b(f)) = \sum_{i=0}^M \lVert\nabla \mu(b(f))|_{\textbf{p}_i}\rVert_{2} + \kappa \sigma (b(f))|_{\textbf{p}_i}
\end{equation}
where $b(f)$ is the belief the robot has over the objective function $f$, $\nabla$ is the gradient operator, $\mu$ and $\sigma$ denote the mean and variance functions respectively, and $\textbf{p}_i = \mathcal{T}(\Theta, \textbf{p})|_{u=i/M}$. From Equation \ref{eq:traj_val_belief}, one can notice parameter $\kappa$ is used to balance exploitation (first term) of high gradient and exploration (second term) of areas of high uncertainty. The exploitation term is designed to encourage the robot to get more samples in quickly varying areas, which often are the most complicated areas to map.

For any predefined set of trajectories, the robot can simulate the value $r$ of each trajectory based on its current pose and belief. The myopic approach to continuous path planning consists of finding the trajectory $r^*$ whose value is maximal:
\begin{equation}
\label{eq:myopic_continuous_planning}
r^*(\textbf{p},b(f)) = \arg \max_{\Theta \in \mathcal{A}} r(\Theta, \textbf{p}, b(f)).
\end{equation}
The method guarantees to find the most informative trajectory based on the chosen acquisition function. In our experiments, we refer to this method as \textit{myopic}. Note that other choices of acquisition function will result in different behaviours. We leave the investigation of the effect of different acquisition functions for future work.
The main shortcoming of the technique presented here is that it is myopic. Indeed, scenarios in which robots need to plan several steps ahead are numerous.

To extend this solution to $n$-look-ahead, we propose to formulate our problem as a POMDP in which the transition and reward functions are assumed to be known, and rewards are given by Equation \ref{eq:traj_val_belief}.

\subsection{A non-myopic approach}
We choose to formulate our problem as a POMDP to take advantage of the framework's previous work and $n$-look-ahead planning capabilities. Using a POMDP instead of a simple Markov Decision Process enables us to encode the fact that the robot takes actions only based on noisy observations of the objective function $f$. The state of the system $\{f, \textbf{p}\}$, fully described by $f$ and the robot's pose $\textbf{p}$, is never completely given to the robot. However, because we are not interested in learning the transitions of the system, the transition function $T$ is explicitly encoded so that the robot can simulate sequences of actions. Similarly, the robot is also given an approximate reward function $\tilde{R}(\{b(f), \textbf{p}\}, a)$ based its current belief $b(f)$ in lieu of $f$. The approximate reward function is based on Equation \ref{eq:traj_val_belief}, encoding the desired exploration-exploitation trade-off.
\\
Let us now define the POMDP formulation of our problem:

\begin{list}{-}{}
\item S: The state is a tuple $\{f, \textbf{p}\}$, where $f$ is the objective function and $\textbf{p}$ is the robot's pose. Note that the robot is never given $f$, but has access to its pose $\textbf{p}$.
\item A: The actions are defined by parameters $\Theta$, each pair of action parameters and pose $\textbf{p}$ fully defining a trajectory $\mathcal{T}(\Theta, \textbf{p})$ starting from $\textbf{p}$. All trajectories are defined over the domain of $f$, and observations are gathered along them.
\item T: The transition function models the probability $T(\{f,\textbf{p}\},\Theta,\{f',\textbf{p'}\})$ of resulting in state $\{f',\textbf{p'}\}$ given trajectory $\mathcal{T}(\Theta, \textbf{p})$ was taken in state $\{f,\textbf{p}\}$.

We assume the transition function to be deterministic, and independent from the objective function $f$ because no transition affects $f$, nor $f$ affects transitions. Thus, it can be rewritten $T(\{\textbf{p}\},\Theta,\{\textbf{p'}\}) = \delta(\mathcal{T}(\Theta,\textbf{p})|_{u=1} - \textbf{p'})$, where $\mathcal{T}(\Theta,\textbf{p})|_{u=1}$ is the resulting pose after executing the full trajectory defined by $\Theta$ and starting from pose $\textbf{p}$, and $\delta$ is the dirac function.
\item R: The reward function computes the sum of rewards obtained along trajectory $\mathcal{T}(\Theta, \textbf{p})$, and is defined as:
\begin{equation}
\label{eq:approx_reward_fn}
\tilde{R}(\{b(f), \textbf{p}\}, \Theta) = r(\Theta, \textbf{p}, b(f)) + cost(\mathcal{T}(\Theta, \textbf{p}))
\end{equation}
where $cost(\mathcal{T}(\Theta, \textbf{p}))$ is the application-specific cost of moving along $\mathcal{T}(\Theta, \textbf{p})$.
\item $\Omega$: The observations are noisy evaluations of $f$ along trajectories.
\item O: At each location, the observation function $O$ is solely defined by the robot's pose. It relies on the distribution $p(o|f(x))$ from which noisy observations of the objective function $f$ are drawn. In the discrete case, $O$ is sampled along trajectories, resulting in a set of observations $\{o_i\}$:
\begin{equation}
\label{eq:observation_fn}
O(\{o_i\},\Theta,\{f,\textbf{p}\})=\prod_{x_i \in \mathcal{T}(\Theta, \textbf{p})}^{} p(o_i|f(x_i)).
\end{equation}
\end{list}

We are interested in methods for finding the optimal policy $\pi^*$, granting the maximum sum of discounted rewards as defined in Equation \ref{eq:policy}. In practice, we need to approximate the infinite sum by a finite one, therefore restricting $\pi^*$ to be the optimal policy over a horizon of $n$ actions.

In the next section, we will present a method for finding an approximation for such policy.

\begin{algorithm}
\caption{MCTS algorithm for BO-POMDP}
\label{alg:MCTS} 
\begin{algorithmic}[1]
\Function{$a^*$ = MCTS}{$b(f), \textbf{p}, depth_{max}$}
	\State $v_0 = NewNode(b(f), \textbf{p}, depth_{max})$
	\For{$i \leftarrow 0, i < $\{Max MCTS iterations\}$, i \leftarrow i+1$}
		\State $v_l \leftarrow TreePolicy(v_0)$
        \State $seq \leftarrow$ $depth_{max} - Depth(v_l)$ random actions
        \State $r \leftarrow$ Simulate $seq$ starting from $v_l$
        \State Back up reward $r$ up the tree.
        \State Update visited counters for $v_l$ branch.
	\EndFor
	\State $v^* \leftarrow$ child of $v_0$ with max accumulated reward
	\State \Return $a^* =$ action from $v_0$ to $v^*$
\EndFunction

\Function{$v_l$ = TreePolicy}{$a$}
	\State $v \leftarrow v_0$
	\While{$Depth(v) \le depth_{max}$}
		\If{$v$ has untried actions}
			\State Choose $a$ from untried actions
			\State $r \leftarrow$ Simulate $a$ and get reward
			\State Update $b(f)$ and \textbf{p}
			\State \Return $v_l = NewNode(b(f),\textbf{p},r)$
		\Else
			\State $v = BestChild(v)$
		\EndIf
	\EndWhile
	\State \Return $v$
\EndFunction

\Function{$v_c$ = BestChild}{$v_p$}
	\State $V \leftarrow$ Children of $v_p$
	\For{$v_i \in V$}
		\State $V \leftarrow$ Visited counter of $v_p$
		\State $N_i \leftarrow$ Visited counter of $v_i$
		\State $R_i \leftarrow$ Accumulated reward of $v_i$
		\State $g(i) = \frac{R_i}{N_i} + \kappa_{MC}\sqrt{\frac{2\ln(N_p)}{N_i}}$
	\EndFor
	\State \Return $\arg\max_{v_i \in V} g(i)$
\EndFunction
\end{algorithmic}
\end{algorithm}

\subsection{Solving continuous non-myopic planning}
\label{sec:nonmyopicSolution}
In BO-POMDP, the belief $b(f)$ the agent holds of the objective function is a probability distribution over the space of functions $f$. As such, we choose to use a Gaussian process to represent it, allowing the belief to potentially capture a very wide variety of objective functions. The update rule of $b(f)$ simply consists in adding a data point $\{\textbf{x}, o\}$ to the GP, where $o$ is a noisy observation at location $\textbf{x}$. Lastly, the mean and variance functions $\mu$ and $\sigma$ required by Equation \ref{eq:traj_val_belief} are naturally provided by the GP.

We use the \textit{Monte-Carlo Tree Search} algorithm to solve the previously formulated POMDP. MCTS has been proven to effectively plan on large discrete POMDPs in \cite{silver2010monte}. MCTS is a partial tree search relying on a metric to balance exploring new branches or extending previous ones. When applied to BO-POMDP, the algorithm searches the space of available sequences of actions at a given step, represented in the form of a tree. Each node $v_i$ of the tree is a fictive state $\{b_i(f), \bm{p_i}\}$, and branches emerging from $v_i$ are actions starting from pose $\bm{p_i}$. Simulating taking an action from node $v_i$ results in the fictive state $\{b_j(f), \bm{p_j}\}$, which is appended to the tree as $v_j$, a child of $v_i$. The tree is grown by simulating actions until a fixed depth while keeping track of simulated rewards obtained in the process. Because not all branches can be explored, MCTS approximates the true expected reward of a leaf node by simulating sequences of random actions from the leaf itself. The algorithm finishes when it reaches a predefined number of iterations, returning the action associated with the maximum expected sum of discounted rewards.

A simplified version of the algorithm is shown in Algorithm \ref{alg:MCTS}. Each iteration comprises four phases. A node is first \textit{selected} according to the exploration metric defined in $BestChild$. The selected node $v_i$ is then \textit{expanded} in function $TreePolicy$, by simulating an untried action starting from its state $\{b_i(f), \bm{p_i}\}$, resulting in node $v_j$. A sequence of random actions is then \textit{simulated} starting from the new leaf node $v_j$, yielding accumulated reward $r$. Finally, reward $r$ is \textit{back-propagated} up the tree, and the branch's visited counters are increased. The algorithm returns the action leading to the first-level child with maximal accumulated reward.

Because this algorithm does not explore the full tree of possible actions, its complexity is not exponential with the planning depth. The most expansive operation being the belief update in line 19, the complexity is only linear with the number of MCTS iterations. As such, one can trade off search accuracy for run time by tuning this parameter.
\\

\section{Experimental results}
\label{sec:ExperimentalResults}
In this section, we demonstrate the behaviour of a robot running BO-POMDP, learning a spatial environmental phenomenon by picking the most informative paths over which to gather samples. We first show key characteristics of our algorithm and compare it to myopic planners in simulator. We then give qualitative results of complex simulated terrain reconstruction and demonstrate the method is fit to run on a real UAV.
\\

So far, actions have been defined to be trajectories $\mathcal{T}(\Theta,\textbf{p})$ parametrised by $\Theta$ and starting from pose $\textbf{p}$. Trajectories are functions of $u \in [0, 1]$ so that $\mathcal{T}(\Theta,\textbf{p})|_{u=0}=\textbf{p}$ and $\mathcal{T}(\Theta,\textbf{p})|_{u=1}=\textbf{p'}$ where $\textbf{p'}$ is the resulting pose after executing the full trajectory. Note that any type of trajectory can be used with this setting.

In our experiments, we choose to define trajectories as 2D cubic splines, functions from $\mathbb{R}$ to $\mathbb{R}^2$, defined as $C(u|\bm{\beta})$  where $u \in [0, 1]$ so that $C = [C_x, C_y]^T = \bm{\beta} [1,u,u^2,u^3]^T$ where $\bm{\beta}$ is a 2-by-4 matrix of parameters fully defining the spline $C$. To ensure continuity from one trajectory to the next, constraints are applied to the parameters $\bm{\beta}$. Restraining the trajectory to start at pose \textbf{p} and have a unit total length is equivalent to:
\begin{align}
\label{eq:splineConst}
C_x|_{u=0}&=\bm{p_x}&\quad
C_y|_{u=0}&=\bm{p_y} \nonumber \\
\\
\frac{\partial C_y / \partial u}{\partial C_x / \partial u} &= \bm{p_\alpha}&\quad
\int_0^1 C(u|\bm{\beta}) du &= 1 \nonumber
\end{align}
The above equations allow for easy generation of a set of splines which can be used as the set of available actions in the POMDP. A set of five discrete actions is generated, allowing the robot to move forward or take slight or sharp turns on both sides. The robot travels at a constant speed and gathers observations at regular time intervals, so that 8 samples are gathered per trajectory. 

The state of the world is composed of the robot's pose $\bm{p}$ and the objective function $f$. The robot is assumed to have access to its pose, but needs to learn a representation of $f$. We use a Gaussian process to maintain a belief over $f$, providing the robot with both point estimate and uncertainty of the objective function. The kernel used varies with the type of environmental variable studied. In our experiments, both the RBF and Mat\'ern kernels are used. Slow changing and smooth variables such as hilly terrains are well represented with the RBF kernel, while Mat\'ern works well harsh changes in the objective function, like sharp objects (e.g. boxes or buildings).
\\

In the following experiments, we address the problem of terrain reconstruction by flying a UAV at constant altitude and taking vertical distance measurements. Because all computation is done in real time, planning many steps ahead is intractable. We therefore restrict our non-myopic algorithms to a maximum look-ahead of 3.

\subsection{Simulated experiments}
\begin{figure}
\centering
\includegraphics[width=80mm]{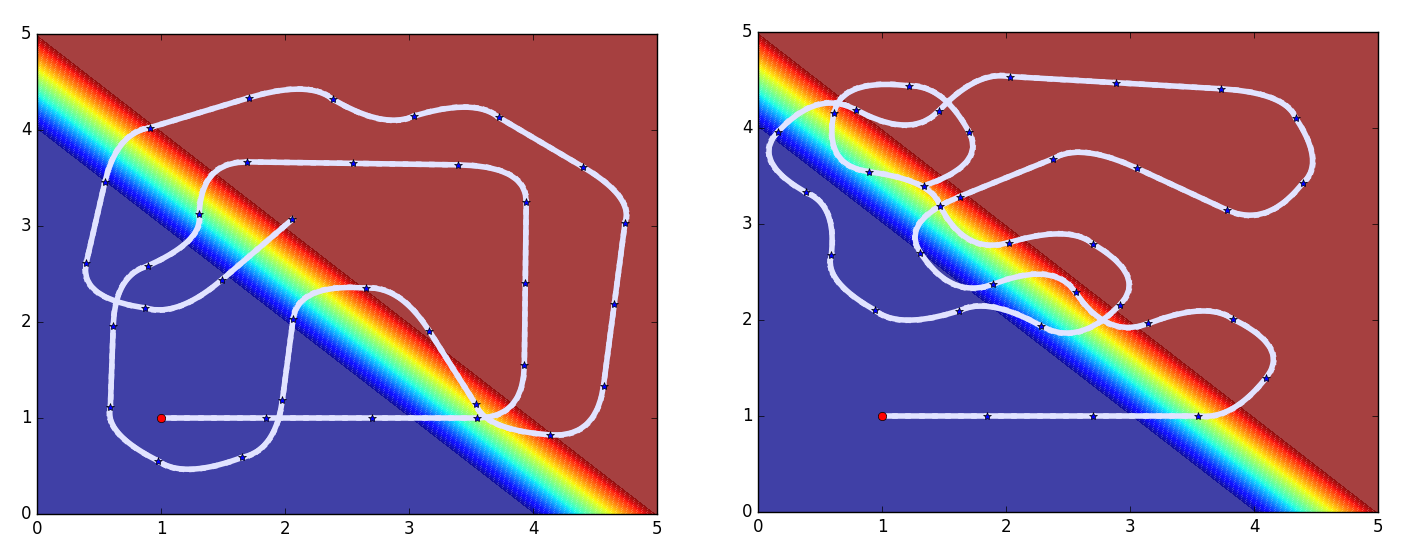}
\caption{Trajectories followed by myopic explorer (left) and our myopic planner (right). The background colours show the monitored objective function, going from low (blue) to high (red) values. The high-gradient diagonal is exploited by our myopic planner.\label{fig:compTraj}}
\end{figure}

The reward function we use in our experiments, defined in Equation \ref{eq:traj_val_belief}, balances exploration behaviour with further sampling of high-gradient areas.
The first experiment aims to demonstrate the behaviour achieved by using such reward function. The \textit{myopic} algorithm presented in Section \ref{sec:myopicSolution} is compared with a \textit{myopic explorer} on a simple environment featuring a high gradient diagonal. The \textit{myopic explorer} favours actions that reduce its belief uncertainty the most, therefore focusing only on exploration. Both algorithms run for 35 steps; their trajectories and the objective function are shown in Figure \ref{fig:compTraj}.
The resulting trajectory shows a clear difference in behaviour, with our myopic planner exploiting high-derivative areas of the objective function. This behaviour directly stems from the reward function used.
\\

\begin{figure}
\centering
\includegraphics[width=40mm]{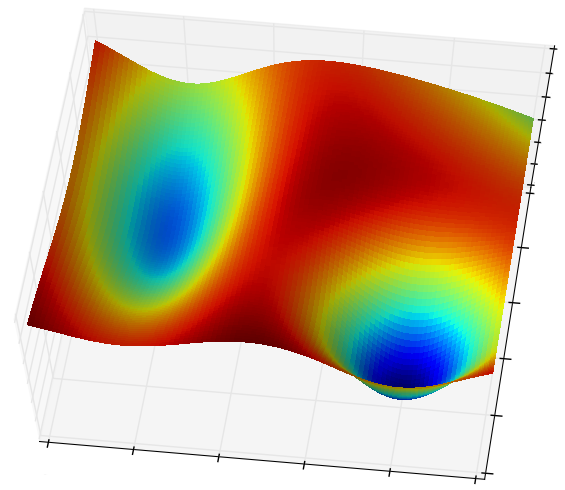}
\caption{Terrain used in second experiment. The pit in the bottom right corner has higher gradient than the top left one. \label{fig:twoPitsTerrain}}
\end{figure}

The second experiment compares the performance of our \textit{myopic} and BO-POMDP algorithms to that of the previous work we build on. The objective function used is shown in Figure \ref{fig:twoPitsTerrain}.
It was designed so that the two pits are separated by a distance which cannot be travelled in a single step, making myopic planners less likely to switch pits.

We first consider the \textit{myopic} method, which searches exhaustively all possible immediate actions. We then examine the BO-POMDP method, which only searches a subset of all sequences of possible actions, limiting its search depth to 3. The method on which we build our work, presented in \cite{marchantRS14}, is denoted as SBO and is also limited to a search depth of 3. A random exploration behaviour was included as baseline. All algorithms are executed for 50 steps, averaged over 50 trials, with an exploration-exploitation parameter $\kappa=5$.

\begin{figure}
\centering
\includegraphics[width=80mm]{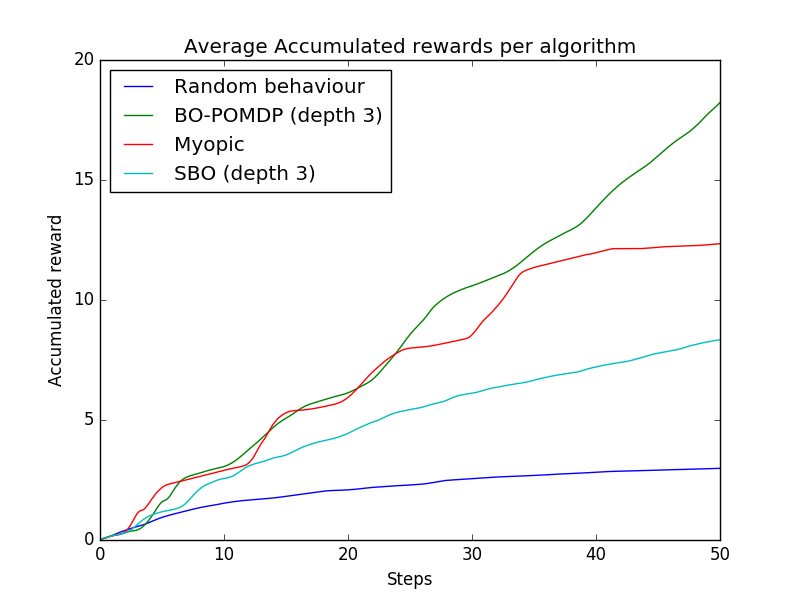}
\caption{Accumulated rewards for second experiment.\label{fig:accRewExp}}
\end{figure}
Figure \ref{fig:accRewExp} compares the accumulated rewards of each method. BO-POMDP shows higher accumulated rewards than its myopic competitor, showing non-myopic methods demonstrate superior behaviour. Indeed, the higher slope of BO-POMDP after step 40 suggests the robot exploits the high-gradient pit more often than the other methods, across all trials. The non-myopic character of BO-POMDP allows it to take low-reward immediate actions that lead to high-reward situations after a few steps. It is the case when the robot switches from the low reward pit (low gradient) to the high reward one (high gradient).

Both of our methods \textit{myopic} and BO-POMDP greatly outperformed SBO. This is because SBO has no incentive to exploit high-gradient areas and therefore performs poorly on the given task.

Several error metrics are used to compare final belief with the ground truth. The following metrics were computed:
\begin{list}{-}{}
\item Root mean square error (RMSE).
\item Weighted root mean square error (WRMSE). Similar to RMSE, weighted so that errors in high-gradient locations are penalised:
\end{list}
\begin{equation}
\label{eq:wrmse}WRMSE = \sqrt[]{\frac{\sum_{i=1}^{N}(\mu(x_i) - f(x_i))^2 \lVert \frac{(\nabla f(x_i) - min \nabla f)}{max \nabla f - min \nabla f}\rVert_{2}}{N}}
\end{equation}
\begin{list}{-}{}
\item Mean Negative Log Likelihood (MNLL). Accounts for the mean value and uncertainty of the final belief:
\end{list}
\begin{equation}
MNLL = \frac{1}{N}\sum_{i=1}^N\frac{1}{2}log(2\pi\sigma(x_i)) + \frac{(\mu(x_i) - f(x_i))^2}{2\sigma(x_i)}
\end{equation}

\begin{table}
\caption{Reconstruction error for second experiment.\label{table:expResults}}
\begin{center}
\begin{tabular}{ l | c | c | c }
  Algorithm & RMSE & WRMSE & MNLL\\
  \hline \hline
  Random & 28.4 & 24.3 & 7.64 \\
  \hline
  SBO & 18.5 & 15.7 & 0.329 \\
  \hline
  Myopic & 8.34 & 7.08 & -1.16 \\
  \hline
  BO-POMDP & 7.92 & 6.7 & 0.623 \\
\end{tabular}
\end{center}
\end{table}
Table \ref{table:expResults} displays final belief errors with respect to ground truth for each method.
\textit{Myopic} or BO-POMDP perform better than random or SBO on all metrics, showing that the reward function reformulation in Equation \ref{eq:traj_val_belief} leads to better mapping quality.
On this simple environment, BO-POMDP outperforms \textit{myopic} on most metrics. This suggests its ability to plan ahead enables it to get measurements in more informative locations, thus leading to better terrain reconstruction. However, \textit{myopic} achieves best performance on MNLL, because its global belief uncertainty is smaller than that of other methods. 
\\

\begin{figure}
\centering
\includegraphics[width=80mm]{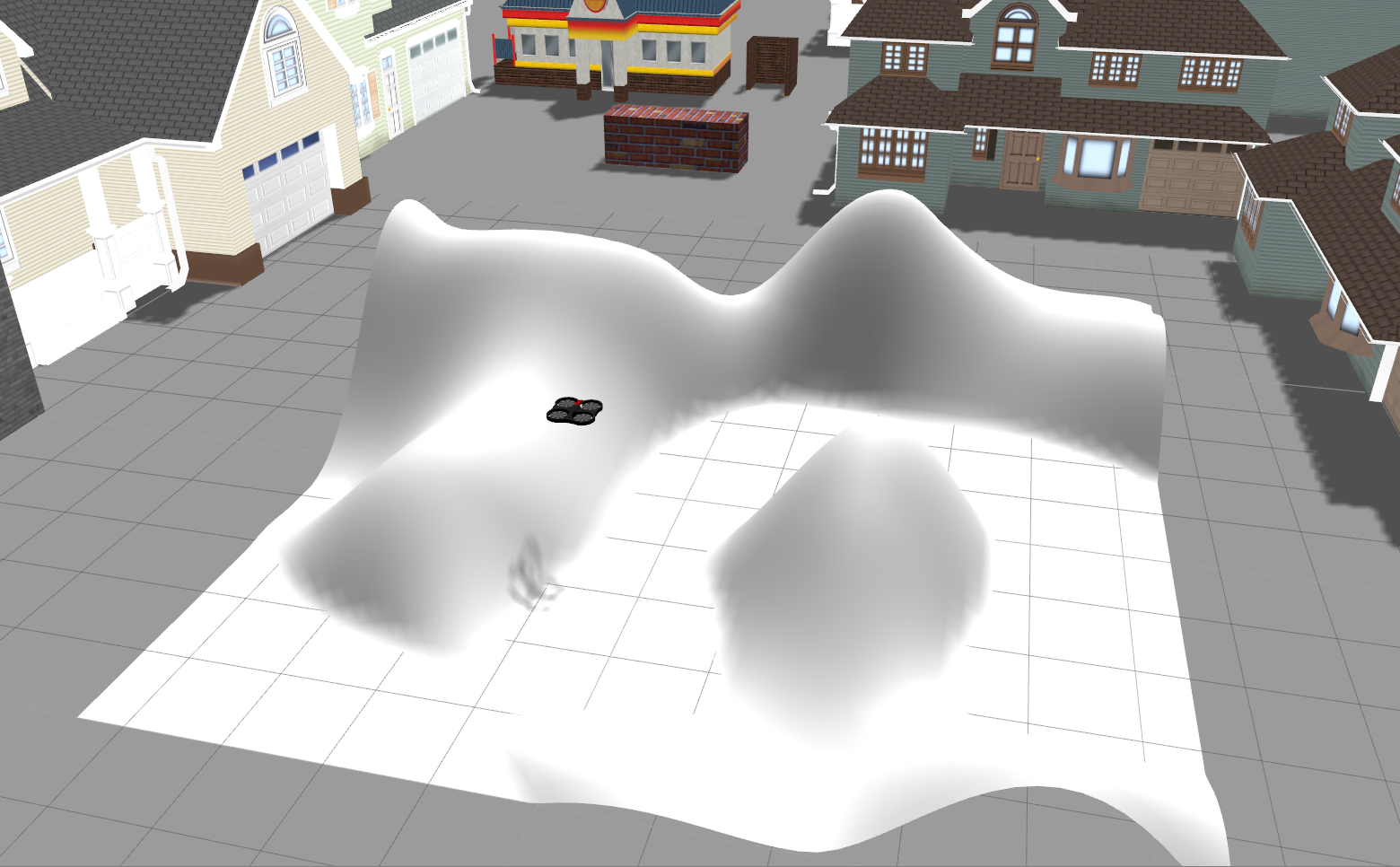}
\caption{Drone flying over a hilly terrain in gazebo simulator. The houses in the background are used for PTAM positioning. \label{fig:experimentArea}}
\end{figure}

\begin{figure}
\centering
\includegraphics[width=80mm]{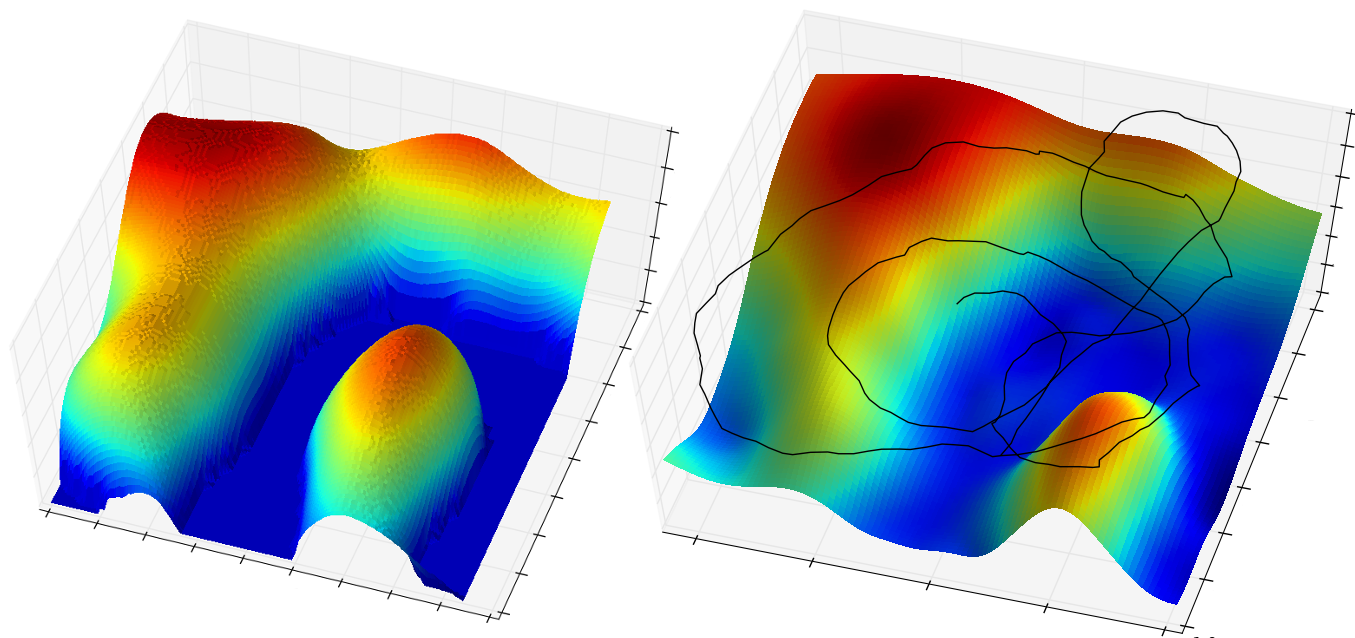}
\caption{Surface reconstruction after simulated experiment with BO-POMDP depth 3 (right), trajectory is displayed in black. Ground truth of the hilly terrain (left), as shown in Figure \ref{fig:experimentArea}.\label{fig:finalBeliefSimulation}}
\end{figure}

Our last simulated experiment is run using \textit{Gazebo} and \textit{tum\_simulator}\footnote{Software available at http://wiki.ros.org/tum\_simulator} to simulate an ARDrone quad-copter. The robot's pose is estimated by running PTAM on the frontal on-board camera. A distance sensor attached underneath the drone gives distance measurement, with respect to the object directly under it. In this experiment, we propose to map a hilly terrain. The experimental setup can be seen in Figure \ref{fig:experimentArea}.
The BO-POMDP algorithm is run for 25 steps, with look-ahead of 3 and $\kappa=10$. Figure \ref{fig:finalBeliefSimulation} shows a 3D terrain reconstructed from in-flight data, compared to ground truth. The reconstructed terrain is very similar to the original one, with only few differences. The final belief uncertainty is high in the four corners, explaining higher error in all corners. Errors could be reduced by running the algorithm for a few additional steps.

\subsection{Real-world experiment}
\begin{figure}
\centering
\includegraphics[width=50mm]{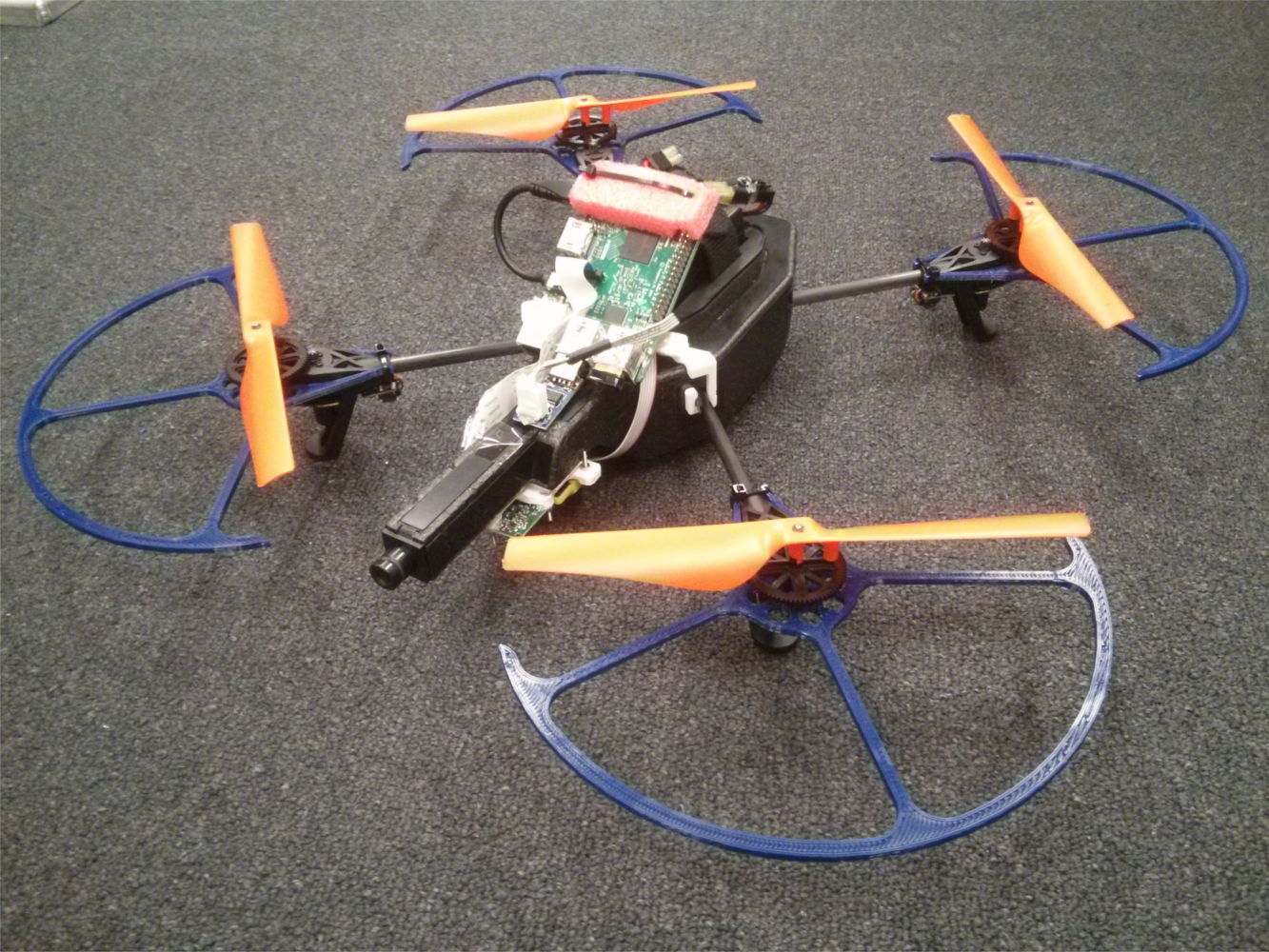}
\caption{Quad-copter used for real-world experiments.\label{fig:ardrone}}
\end{figure}

\begin{figure}
\centering
\includegraphics[width=80mm]{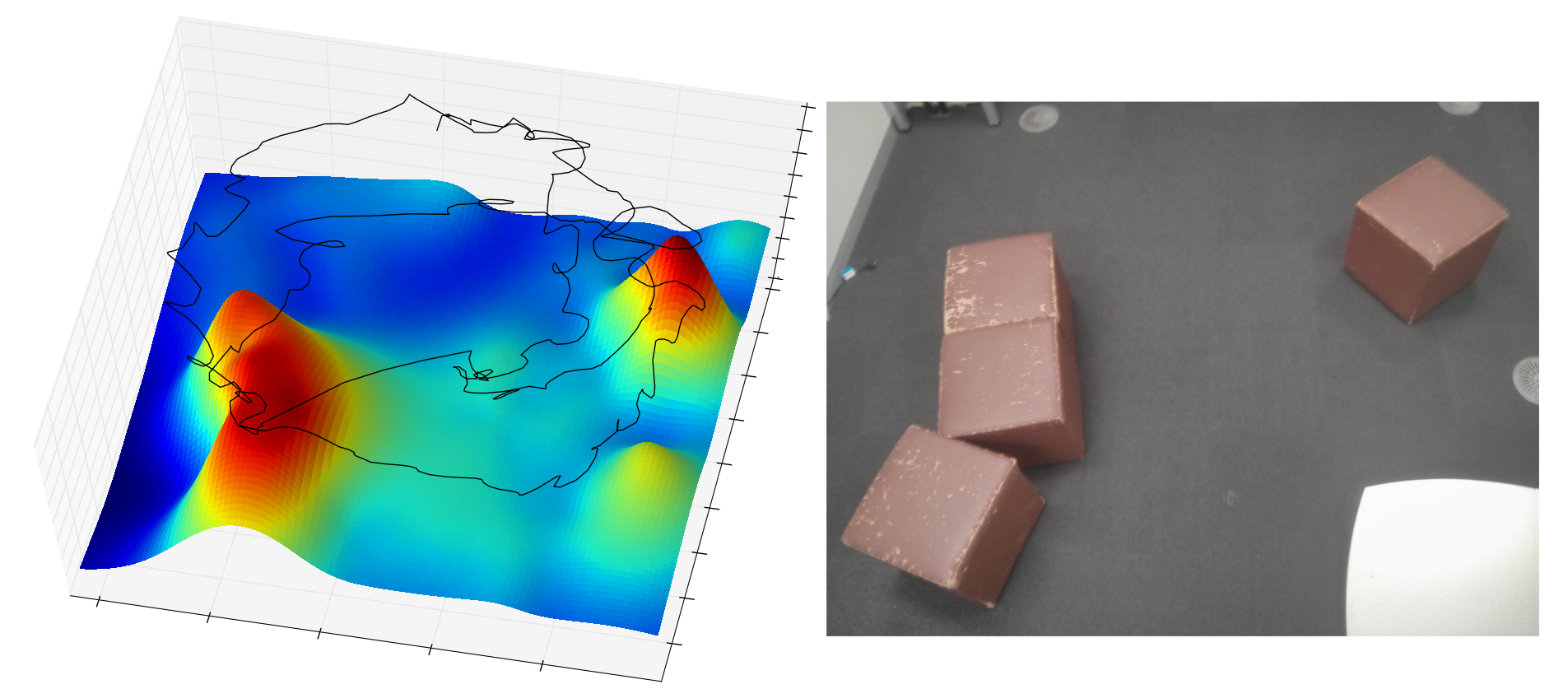}
\caption{Surface reconstruction after real-world experiment with BO-POMDP depth 3 (left), trajectory displayed in black. Photo of the corresponding 3x3 meter experiment area (right). \label{fig:finalBeliefReal}}
\end{figure}

\begin{figure}
\centering
\includegraphics[width=50mm]{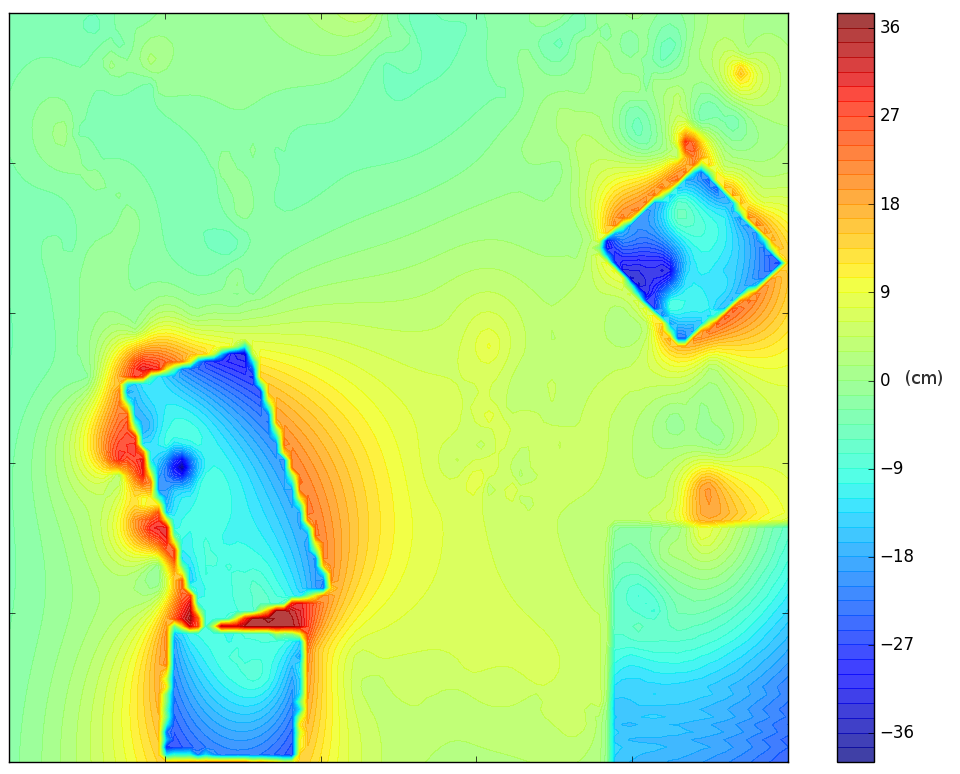}
\caption{Centred surface reconstruction error. A GP with Exponential kernel was used to reconstruct the map from flight data. The highest error is achieved around prop edges. \label{fig:reconstructionErr}}
\end{figure}

For the last experiment, an ARDrone quad-copter was used. It was modified to include an infra-red range sensor facing down and a Raspberry Pi mounted on top, running ROS and communicating with a base station. See Figure \ref{fig:ardrone}.
Similarly to the previous experiments, the goal is to reconstruct the 3D surface of an area. The environment of study features an 3x3 meter indoor area with sharp-edged props, as displayed in Figure \ref{fig:finalBeliefReal} (right).
The robot's pose is estimated by running the same PTAM algorithm as in the previous experiment. All computation runs in real time, most of it in a single thread on a base station featuring a 2.90GHz i7 CPU. Because of real-time constraints and flight time limitations, planning was only carried out with depth 3, and the number of steps restricted to 15. In this configuration, choosing the next trajectory takes less than 5 seconds for each step. The 3D surface reconstructed from data gathered during the experiment is shown in Figure \ref{fig:finalBeliefReal} (left). While one can clearly see the resemblance between the reconstructed surface and the real environment, reconstruction is not perfect. All props were detected by the range sensor, and positioned properly. However, their sharp edges are not well represented in the 3D surface, and notably the hole between the props on the left side of the area. The reconstruction error in Figure \ref{fig:reconstructionErr} confirms edges are where error is the highest. This problem is likely due to a lack of data, resulting from short flight time. Note that the belief uncertainty is high in several locations after the flight finished, meaning the belief could be further improved with more exploration.
The range sensor angle (3 degrees) also contributes to smoothing edges. Using a laser sensor instead would yield more precise and much sharper measurements.
The mean and variance of the reconstruction error are $6.01$ and $1.98$ cm respectively. The mean value can be disregarded as it only is an offset. However, the error variance is small in comparison to the sensor's theoretical accuracy of $\pm 4 cm$, suggesting low sensor accuracy deteriorated mapping quality.

Using a sharper GP kernel such as the exponential kernel improves edge reconstruction, but fails in areas where data is sparse. Running the experiment for longer would allow more data to be gathered and probably lead to more accurate terrain reconstruction with such kernel.

Despite the fact our last experiment suffered from practical limitations, it clearly shows our method can be implemented and run on real UAVs, and successfully applied to real life scenarios of environment monitoring.

\section{Conclusion}
\label{sec:Conclusion}
In this paper we present BO-POMDP, a non-myopic planner relying on a POMDP formulation of sequential BO, first introduced in \cite{marchantRS14}. Our first contribution is the reformulation of the reward function to balance exploitation of high-gradient areas and exploration. We show in simulated experiments how the behaviour emerging from the new reward function favours varying areas over constant ones, yielding better mapping accuracy and accumulated reward than previous methods. Comparisons of 3D terrain reconstruction shows BO-POMDP achieves lower errors than its myopic equivalent.

Our second contribution is the adaptation of our method to real-world monitoring problems. We first showed promising results of terrain reconstruction in a realistic simulator. Our method was then deployed on a cheap quad-copter with very limited sensing capabilities to achieve decent mapping of an indoor area.
While real-world experiments did not yield a very accurate reconstructed map, it should be noted that the problem of accurate edge reconstruction is challenging when dealing with so few and sparse data. Reconstruction could be greatly improved by upgrading the single-beam infra-red sensor to a more accurate multi-beam laser sensor, and increasing flight time.

We believe that choosing trajectories that exploit high gradients leads to more informative data gathering, thus improving mapping accuracy. The method we described can be applied to a wide variety of environmental phenomena to improve mapping quality.
As future work, we plan to address the problem of environmental monitoring when discretization of the action space is impossible.

\bibliographystyle{IEEEtran.bst}
\bibliography{references}
\end{document}